# Accelerating supply chains with Ant Colony Optimization across range of hardware solutions

Ivars Dzalbs, Tatiana Kalganova[1]

[1] Brunel University London, Kingston Lane, Uxbridge, UB8 2PX, UK
Tatiana.Kalganova@brunel.ac.uk

**Abstract.** Ant Colony algorithm has been applied to various optimization problems, however most of the previous work on scaling and parallelism focuses on Travelling Salesman Problems (TSPs). Although, useful for benchmarks and new idea comparison, the algorithmic dynamics does not always transfer to complex real-life problems, where additional meta-data is required during solution construction. This paper looks at real-life outbound supply chain problem using Ant Colony Optimization (ACO) and its scaling dynamics with two parallel ACO architectures – Independent Ant Colonies (IAC) and Parallel Ants (PA). Results showed that PA was able to reach a higher solution quality in fewer iterations as the number of parallel instances increased. Furthermore, speed performance was measured across three different hardware solutions – 16 core CPU, 68 core Xeon Phi and up to 4 Geforce GPUs. State of the art, ACO vectorization techniques such as SS-Roulette were implemented using C++ and CUDA. Although excellent for TSP, it was concluded that for the given supply chain problem GPUs are not suitable due to meta-data access footprint required. Furthermore, compared to their sequential counterpart, vectorized CPU AVX2 implementation achieved 25.4x speedup on CPU while Xeon Phi with its AVX512 instruction set reached 148x on PA with Vectorized (PAwV). PAwV is therefore able to scale at least up to 1024 parallel instances on the supply chain network problem solved.

## 1. Introduction and motivation

Supply chain optimization has become an integral part of any global company with a complex manufacturing and distribution network. For many companies, inefficient distribution plan can make a significant difference to the bottom line. Modelling a complete distribution network from the initial materials to the delivery to the customer is very computationally intensive. With increasing supply chain modelling complexity in ever changing global geo-political environment, fast adoptability is an edge. A company can model impact of currency exchange rate changes, import tax policy reforms, oil price fluctuations and political events such as Brexit before they happen. This requires fast optimization algorithms.

Mixed Integer Linear Programming (MILP) tools such as Cplex are commonly used to optimize various supply chain networks [1]. Although MILP tools are able to obtain optimum solution for large variety of linear models, not all real-world supply chain models are linear. Furthermore, MILP is computationally expensive and on large instances can fail to produce an optimal solution. For that reason, many alterative algorithmic approaches (heuristics, meta-heuristics, fuzzy methods) have been explored to solve large-complex SC models [1]. One of these algorithms is the Ant Colony Optimization (ACO), which can be well mapped to real world problems such as routing [2] and scheduling [3]. Supply Chain Optimization Problem (SCOP) includes both, finding the best route to ship a specific order and finding the most optimal time to ship it, such that it reaches expected customer satisfaction while minimizing the total cost occurred.

Ant colony algorithms try to mimic the observed behavior of ants inside colonies, in order to solve a large range of optimization problems. Since the introduction by Marco Dorigo in 1992, many variations and hybrid approaches of Ant Colony algorithms have been explored [4] [5]. Most ant colony algorithms consist of two distinct stages – solution construction and pheromone feedback to other ants. Typically, an artificial ant builds its solution from the pheromone left from previous ants, therefore allowing communication over many iterations via a *pheromone matrix* and converges to a better solution. The process of solution creation and pheromone update is repeated over many iterations until the termination criterion is reached, this can be either total number of iterations, total computation time or dynamic termination.

Researchers in [6] compared an industrial optimization-based tool – IBM ILOG Cplex with their proposed ACO algorithm. It was concluded that the proposed algorithm covered 94% of optimal solutions on small problems and 88% for large-size problems while consuming significantly less computation time. Similarly, [7] compared ACO and Cplex performance on multi-product and multi-period Inventory Routing Problem. On small instances ACO reached 95% of optimal solution while on large instances performed better than time-constrained Cplex solver. Furthermore, ACO implementations of Closed-Loop Supply Chain (CLSC) have been proposed; CLSC contains two parts of the supply chain – forward supply and reverse/return. [8] solved CLSC models, where the ACO implementation outperformed commercial MILP (Cplex) on nonlinear instances and obtained 98% optimal solution with 40% less computation time on linear instances.

The aim of this paper is to explore parallelism techniques across multiple hardware solutions for a real-world supply chain optimization problem (where meta-data overhead during solution construction plays a significant role on the total compute time). The paper is structured as follows: Section 2 explores current state of the art parallel implementations of ACO across CPU, GPU and Xeon Phi; Section 3 introduces the hardware and software solutions used; Section 4 described the real-world problem being solved; Section 5 details the parallel ACO implementations and Section 6 compares the results. Finally, Section 7 concludes the paper.

## 2. Parallel Ant Colony Optimization

Since the introduction of ACO in 1992, countless ACO algorithms have been applied to many different problems and many different parallel architectures have been explored previously. [9] specifies 5 of such architectures:

- <u>Parallel Independent Ant Colonies</u> – each ant colony develop their own solutions in parallel without any communication in-between;
- <u>Parallel Interacting Ant Colonies</u> – each colony creates solution in parallel and some information is shared between the colonies;
- <u>Parallel Ants</u> – each ant creates solution independently, then all the resulting pheromones are shared for the next iteration;
- <u>Parallel Evaluation of Solution Elements</u> – for problems where fitness function calculations take considerably more time than the solution creation;
- <u>Parallel Combination of Ants and Evaluation of Solution Elements</u> – a combination of any of the above.

Researchers have tried to exploit the parallelism offered from recent multi core CPUs [10], along with clusters of CPUs ( [11] [12]) and most recently GPUs [13] and Intel's many core architectures such as Xeon Phi [14]. Breakdown of the strategies and problems solved are shown in Table 1.

*Table 1. ACO architecture and hardware configurations explored. LAC - Longest Common Subsequence Problem, MKP - Multidimensional Knapsack Problem, TSP - Travelling Salesman problem. IAC – Independent Ant Colonies, IntAC – Interactive Ant Colonies, PA – Parallel Ants.*

|  | Task parallelism, IAC | Task parallelism, IntAC | Task parallelism, PA | Data parallelism, PA |
|---|---|---|---|---|
| CPU | Scheduling [15] | Scheduling [15] | TSP [16] [17] Scheduling [15] **Supply chain [this paper]** | TSP [18] [19] **Supply chain [this paper]** |
| GPU | n/a | n/a | Protein folding [20] TSP [16] MKP [21] LAC [22] | TSP [23] [24] [18] [19] Edge detection [25] **Supply chain [this paper]** |
| CPU cluster | Scheduling [26] | TSP [9] | TSP [12] | n/a |
| Xeon Phi | n/a | n/a | **Supply chain [this paper]** | TSP [27] [28] [29] **Supply chain [this paper]** |

During the search, an Ant has to keep track of the existing state meta-data, for instance Travelling Salesman Problem only need to keep the track of what cities have been visited as part of problem constraint. However, real-life problems have a lot more constraints and therefore requires a lot of meta-data storage during solution creation. This paper explores such problem in the supply chain domain. Table 2 shows the most common problems solved by ACO and their corresponding associated constraints / meta-data required during solution creation.

*Table 2. Meta-data required during solution creation based on problem type*

| Problem | Meta-data required during solution creation | Comment |
|---|---|---|
| Scheduling | 2 | Resource and precedence constraints |
| TSP | 1 | Has the city been visited |
| Protein Folding | 1 | Has the sequence been visited |
| MKP | 1 | Total weight per knapsack |
| LAC | 1 | Tracking of current position in string |
| Edge detection | 1 | Has edge already been visited |
| **Supply chain (this paper)** | **3** | **Capacity, daily order, freight weight constraints** |

## 2.1. CPU

Parallel ACO CPU architectures have been applied to various tasks – for example, [15] applied ACO for mining supply chain scheduling problem. Authors managed to reduce the execution time from one hour (serial) to around 7 minutes. Both [30] and [31] used ACO for image edge detection with varying results, [30] achieved a speedup of 3-5 times while [31] managed to reduce sequential runtime by 30%. Most commonly, ACO has been applied to Travelling Salesman Problem (TSP) benchmarks. For instance, [17] proposed ACO approach with randomly synchronized ants, the approach showed a faster convergence compared to other TSP approaches. Moreover, authors in [19] proposed new multi-core SIMD model for solving TSPs. Similarly, both [32] and [33] tries to solve large instances of TSP (up to 200k and 20k cities respectively) where the architectures are limited to the size of pheromone matrix. [33] discusses such limitations and proposes a new pheromone sharing for local search – effective heuristics ACO (ESACO), which was able to compute TSP instances of 20k. In contrast, authors in [32] eliminate the need of pheromone matrix and store only the best solutions similarly to the Population ACO. Furthermore, researchers implement a Partial Ant, also known as cunning ant, where ant takes existing partial solution and builds on top of it. Speedups of as much as 1200x are achieved compared to sequential Population ACO.

Generally, CPU parallel architecture implementations come down to three programming approaches - Message Passing Interface (MPI) parallelism, OpenMP parallelism [34] and data parallelism with the vectorization of Single Instruction Multiple Data (SIMD). For instance, [35] explored both master-slave and coarse-grained strategies for ACO parallelization using Message Passing Interface (MPI). It was concluded that fine-grained master-slave strategy performed the best. [36] used MPI with ACO to accelerate Maximum Weight Clique Problem (MWCP). Proposed algorithm was comparable to the ones in literature and outperformed Cplex solver in both – time and performance. Moreover, authors in [26] implemented parallel ACO for solving Flow shop scheduling problem with restrictions using MPI. Compared to sequential version of the algorithm, 93 node cluster achieved a speedup of 16x. [37] compared ACO parallel implementation on MPI and OpenMP on small vector estimation problem. It was found that maximum speedup of OpenMP was 24x while MPI – 16x. Furthermore, [18] explored multi-core SIMD CPU with OpenCL and compared it to the performance of GPU. It was found optimized parallel CPU-SIMD version can achieve similar solution quality and computation time than the state of art GPU implementation solving TSP.

## 2.2. Xeon Phi

Intel's Xeon Phi Many Integrated Core (MIC) architecture offers many cores on the CPU (60-72 cores per node) while offering lower clock frequency. Few researchers have had the opportunity to research ACO on the Xeon Phi architecture. For instance, [27] showed how utilizing L1 and L2 cache on Xeon Phi coprocessor allowed a speedup of 42x solving TSP compared to a sequential execution. Due to the nature of SIMD features such as AVX-512 on Xeon Phi, researchers in both [29] and [28] proposed a vectorization model for roulette wheel selection in TSP. In case of [29] a 16.6x speedup was achieved compared to a sequential execution. To the best of authors knowledge, Xeon Phi and ACO parallelism has not been explored to any other problem except TSP.

## 2.3. GPUs

General Purpose GPU (GPGPU) programming is a growing field in computer science and machine learning. Many researchers have tried exploiting latest GPU architectures in order to speed optimize the convergence of ACO. ACO GPU implementation expands to many fields, such as edge detection ( [25] [38]), protein folding [20], solving Multidimensional Knapsack Problems (MKPs) [21] and Vertex coloring problems [39]. Moreover, researchers have used GPU implementations of ACO for classification ( [40] [41]) and scheduling ( [42] [43]) with various speedups compared to the sequential execution. However, majority of publications are solving Travelling Salesman Problems [44], although useful for benchmarking and comparison, little characteristics transfer to other application areas. For instance, highly optimized local memory on GPU (Compute Unified Device Architecture - CUDA) can significantly speed up the execution for TSP, however, when applied to real-life problems where additional restrictions and metadata is required to build a solution, most of the data needs to be stored on much slower global memory. Authors in [16] did extensive research comparing server, desktop and laptop hardware solving TSP instances on both CUDA and OpenCL. Although there are couple ACO OpenCL implementations on GPU ( [45] [22]), the majority of implementations use CUDA. For instance, [46] implemented a GPU-based ACO and achieved a speedup of 40x compared to sequential ACS. Similarly, a 22x speedup was achieved in [47] solving pr1002 TSP and 44x on fnl4461 TSP instance in [48]. However, there are also various hybrid approaches for solving TSP - [49] solves parallel Cultural ACO (pCACO) (a hybrid of genetic algorithm and ACO). Research showed that pCACO outperformed sequential and parallel ACO implementations in terms of solution quality. Furthermore, [50] solved TSP instances using ACO-PSO hybrid and authors in [51] explored heterogenous computing with multiple GPU architectures for TSP.

Although task parallelism has potential for a speedup, [23] showed how data parallelism (vectorization) on GPU can achieve better performance by proposed Independent Roulette wheel (I-Roulette). Authors then expanded the I-Roulette implementation in [24], where SS-Roulette wheel was proposed. Further, [52] implements a G-Roulette – a grouped roulette wheel selection based on I-Roulette, where cities in TSP selection is grouped in CUDA warps. An impressive speedup of 172x was achieved compared to the sequential counterpart.

## 2.4. Comparing hardware performances

Comparing parallel performances of different hardware architectures fairly is by no means trivial. Most research compares a sequential CPU ACO implementation to the one of the parallel GPUs, which is hardly fair [53]. To amplify the issue, unoptimized sequential code is compared to highly optimized GPU code. This results in misleading and inflated speedups [13]. Furthermore, [22] argues that often the parameter settings chosen for the sequential implementation is biased in favor of GPU. [13] proposes a criteria to calculate the real-world efficiency of two different hardware architectures by comparing the theoretical peak performances of GPU and CPU. While the proposed method is more appropriate, it still doesn't account for real-life scenarios where memory latency/speed, cache size, compilers and operating systems all play a role of the final execution time. Therefore, two different systems with similar theoretical floating-point operations per second running the same executable can have significantly different execution times. Furthermore, in some instances only execution time or solution quality is compared, rarely both are taken into consideration when comparing results.

# 3. Background

This section briefly covers the tools and hardware specific languages used in the implementation.

## 3.1. Parallel processing with OpenMP

OpenMP is set of directives to a compiler that allows programmer to create parallel tasks as well as vectorization (Single Instruction Multiple Data - SIMD) in order to speed up execution of a program. Program containing parallel OpenMP directives starts as single thread, when directive such as *#pragma omp parallel* is reached, main thread will create a thread pool and all methods within pragma region will be executed in parallel by each thread in the thread group. Once the thread reaches the end of the region, it will wait for all other threads to finish before dissolving the thread group and only the main thread will continue.

Furthermore, OpenMP also supports nesting, meaning a thread in a thread-group can create its own individual thread-group and become the master thread for the newly created thread-group. However, thread-group creation and elimination can have significant overhead and therefore, thread-group re-use is highly recommended [54].

This paper utilizes both *omp parallel* and *omp simd* directives.

## 3.2. CUDA programming model

Compute Unified Device Architecture (CUDA) is a General-purpose computing model on GPU developed by Nvidia in 2006. Since then this proprietary framework has been utilized in the high-performance computing space via multiple Artificial Intelligence (AI) and Machine Learning (ML) interfaces and libraries/APIs. CUDA allows to write C programs that takes advantage of any recent Nvidia GPU found in laptops, workstations and data centers.

Each GPU contains multiple Streaming Multiprocessors (SM) that are designed to execute hundreds of threads concurrently. In order to achieve that, CUDA implements SIMT (Single Instruction Multiple-Threads) architecture, where instructions are pipelined for instruction level parallelism. Threads are grouped in sets of 32 – called *warps*. Each warp executes one instruction at a time on each thread. Furthermore, CUDA threads can access multiple memory spaces – global memory (large size, slower), texture memory (read only), shared memory (shared across threads in the same SM, lower latency) and local memory (limited set of registers within each thread, fastest) [55].

A batch of threads are grouped into a *thread-block*. Multiple thread-blocks create a *grid of thread blocks*. Programmer specifies the grid dimensionality at kernel launch time, by providing the number of thread-blocks and the number of threads per thread-block. Kernel launch fails if the program exceeds the hardware resource boundaries.

## 3.3. Xeon Phi Knights Landing architecture

Knights Landing is a product code name for Intel's second-generation Intel Xeon Phi processors. First generation of Xeon Phi, named Knights Corner, was a PCI-e coprocessor card based on many Intel Atom processor cores and support for Vector Processing Units (VPUs). The main advancement over Knights Corner was the standalone processor that can boot stock operating systems, along with improved power efficiency and vector performance. Furthermore, it also introduced a new high bandwidth MCDRAM memory. Xeon phi support for standard x86 and x86-64 instructions, allows majority CPU compiled binaries to run without any modification. Moreover, support for 512-bit Advanced Vector Extensions (AVX-512) allows high throughput vector manipulations.

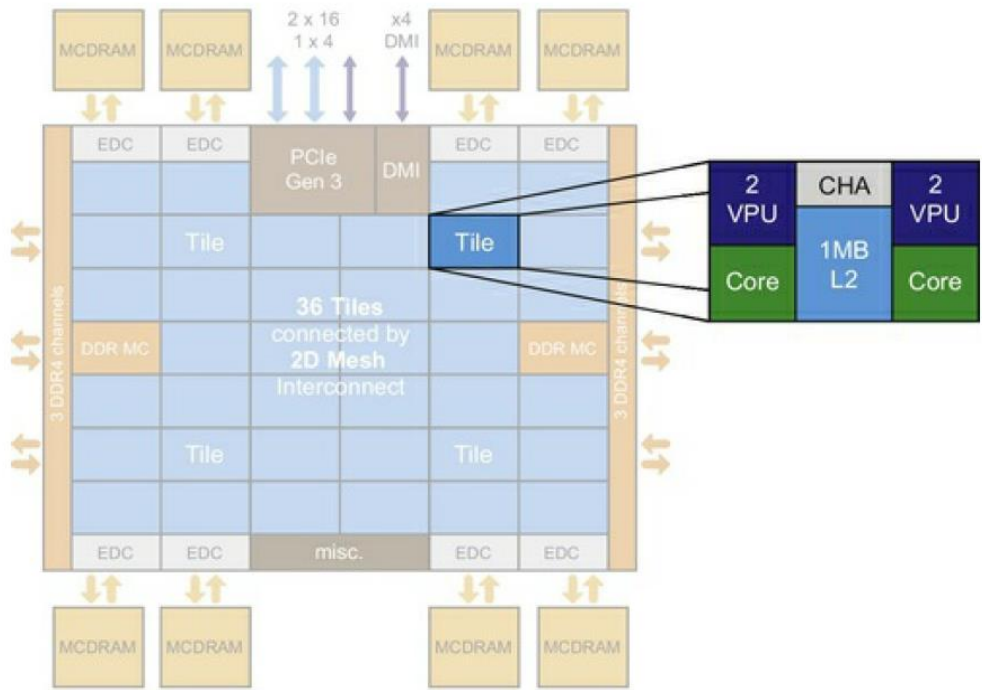

*Figure 1. Knights Landing tile with larger processor die [56]*

The Knights Landing cores are divided into tiles (typically between 32 and 36 tiles in total). Each tile contains two processor cores and each core is connected to two vector processing units (VPUs). Utilizing AVX-512 and two VPUs, each core can deliver 32 dual-precision (DP) or 64 single-precision (SP) operations each cycle [56]. Furthermore, each individual core supports up to four threads of execution – hyper threads where instructions are pipelined.

Another introduction with the Knights Landing is the cluster modes and MCDRAM/DRAM management. Processor offers three primary cluster modes – All to all mode, Quadrant mode and Sub-Numa Cluster (SNC) mode and three memory modes – cache mode, flat mode and hybrid mode. For detailed description of the Knights Landing Xeon Phi architecture refer to [56].

## 4. Problem description

A real-world dataset of an outbound logistics network is provided by a global microchip producer. The company provided demand data for 9216 orders that need to be routed via their outbound supply chain network of 15 warehouses, 11 origin ports and 1 destination port (see Figure 2). Warehouses are limited to a specific set of products that they stock, furthermore, some warehouses are dedicated for supporting only a particular set of customers. Moreover, warehouses are limited by the number of orders that can be processed in a single day. A customer making an order decides what sort of service level they require – DTD (Door to Door), DTP (Door to Port) or CRF (Customer Referred Freight). In case of CRF, customer arranges the freight and company only incurs the warehouse cost. In most instances, an order can be shipped via one of 9 couriers offering different rates for different weight bands and service levels. Although the majority of the shipments are done via air transport, some orders are shipped via ground – by trucks. The majority of couriers offer discounted rates as the total shipping weight increases based on different weight bands. However, a minimum charge for shipment still applies. Furthermore, faster shipping tends to be more expensive, but offer better customer satisfaction. Customer service level is out of the scope of this research.

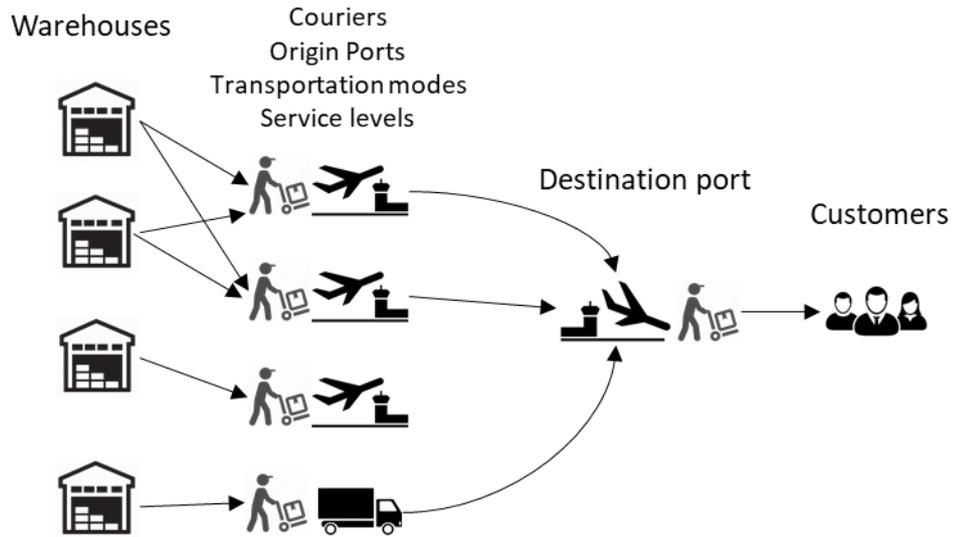

Figure 2 - Graphical representation of the outbound supply chain

### 4.1. Dataset

Dataset [57] is divided into 7 tables, one table for all orders that needs to be assigned a route – *OrderList* table, and 6 additional files specifying the problem and restrictions. For instance, the *FreightRates* table describes all available couriers, the weight gaps for each individual lane and rates associated. The *PlantPorts* table describes the allowed links between the warehouses and shipping ports in real world. Furthermore, the *ProductsPerPlant* table lists all supported warehouse-product combinations. The V*miCustomers* lists all special cases, where warehouse is only allowed to support specific customer, while any other non-listed warehouse can supply any customer. Moreover, the *WhCapacities* lists warehouse capacities measured in number of orders per day and the *WhCosts* specifies the cost associated in storing the products in given warehouse measured in dollars per unit.

### 4.2. Fitness function

The main goal of optimization is to find a set of warehouses, shipping lanes and couriers to use for the most cost-effective supply chain. Therefore the fitness function is derived from two incurred costs – warehouse cost $wc_{ki}$ and transportation cost $tc_{kpj}$ in equation (1). The totaling cost is then calculated across all orders $k$ in the dataset.

$$\min \sum_{k=1}^{l} (wc_{ki} + tc_{kpj}) \qquad (1)$$

Where $wc_{ki}$ is warehouse cost for order $k$ at warehouse $i$ and $tc_{kpj}$ is transportation cost for order $k$ between warehouse port $p$ and customer port $j$, total number of orders $l$.

$$wc_{ki} = q_k \times P_i \tag{2}$$

Where warehouse cost $wc_{ki}$ for order $k$ at warehouse $i$ is calculated in (2), by the number of units in order $q_k$ multiplied by the warehouse storage rate $P_i$ (*WhCosts* table).

1. **if** $s_k = CRF$ **then** $tc_{kpj} = 0$
2. **else if** $m = GROUND$ **then** $tc_{kpj} = \frac{R_{pjcstm}}{\sum_{k=1}^{l} w_{kpjcstm}} \times w_{kpjcstm}$
3. **else if** $R_{pjcstm} \times w_{kpjcstm} < M_{pjcstm}$ **then** $tc_{kpj} = M_{pjcstm}$
4. **else** $tc_{kpj} = R_{pjcstm} \times w_{kpjcstm}$

*Figure 3. Pseudo code for calculating order transportation cost*

Where $s$ is the service level for order $k$. $M_{pjcstm}$ is the minimum charge for given line $pjcstm$, $w_{kpjcstm}$ is the weight in kilograms for order $k$ shipped from warehouse port $p$ to customer port $j$ via courier $c$ using service level $s$, delivery time $t$ and transportation mode $m$. $R_{pjcstm}$ is the freight rate (dollars per kilogram) for given weight gap based on total weight for the line $pjcstm$ (*FreightRates* table).

Furthermore, transportation cost for a given order and chosen line is calculated by algorithm in Figure 3. The algorithm first check what kind of service level the order requires, if the service level is equal to CRF (Customer Referred Freight) – transportation cost is 0. Furthermore, if order transportation mode is equal to GROUND (order transported via truck), order transportation cost is proportional to the weight consumed by the order ($w_{kpjcstm}$) in respect of the total weight for given line $pjcstm$ and the rate charged by courier for full track $R_{pjcstm}$. Moreover, a minimum charge of $M_{pjcstm}$ is applied in cases where the transportation cost is less than the minimum charge. In all other cases, the transportation cost is calculated based on order weight $w_{kpjcstm}$ and the freight rate $R_{pjcstm}$. The freight rate is determined based on total weight on any given line $pjcstm$ and the corresponding weight band in the freight rate table.

### 4.3. Restrictions

Problem being solved complies with the following constraints:

$$\sum_{k=1}^{l} o_{ki} \leq C_i \tag{3}$$

Where $o_{ki} = 1$ if order $k$ was shipped from warehouse $i$ and 0 otherwise. $C_i$ is the order limit per day for warehouse $i$ (*WhCapacities* table).

$$\sum_{k=1}^{l} w_{kpjcstm} \leq \max F_{pjcstm} \qquad (4)$$

Where $w_{kpjcst}$ is the weight in kilograms for order $k$ shipped from warehouse port $p$ to customer port $j$ via courier $c$ using service level $s$, delivery time $t$ and transportation mode $m$. $F_{pjcstm}$ is the upper weight gap limit for line $pjcstm$ (*FreightRates* table).

$$k_z \in i_z \qquad (5)$$

Where product $z$ for order $k$ belongs to supported products at warehouse $i$ (*ProductsPerPlant* table). Warehouses can only support given customer in the *VmiCustomers* table, while all other warehouses that are not in the table can supply any customer. Moreover, warehouse can only ship orders via supported origin port, defined in *PlantPorts* table.

## 5. Methods and implementation

In order to solve the transportation network optimization problem, we are using an Ant Colony System algorithm first proposed by [58]. Because ACO is an iterative algorithm, it does require sequential execution. Therefore, the most naïve approach for parallel ACO is running multiple Independent Ant Colonies (IAC) with a unique seed for the pseudo random number generator for each colony (high level pseudo code in Figure 4). Due to the stochastic nature of solution creation, it is therefore more probabilistic to reach a better solution than a single colony. This approach has the advantage of low overhead as it requires no synchronization between the parallel instances during search. At the very end of the search, the best solution of all parallel colonies is chosen as the final solution. Main disadvantage of IAC is that if one of the colonies finds a better solution, there is no way to improve all the other colony's fitness values.

**Independent Ant Colonies (IAC)**

1. **for** all parallel instances *m* **parallel do**
2.    **for** all iterations *i* **do**
3.       **for** all local ants *a* **do**
4.          local pheromone = global pheromone
5.          construct solution
6.          local pheromone update
7.       **end for**
8.       update global pheromone update based on best solution
9.    **end for**
10. **end for**
11. find best solution across parallel instances

*Figure 4. High level pseudo code for Independent Ant Colonies (IAC) search algorithm*

Alternatively, the ACO search algorithm could also be letting the artificial ant colonies synchronize after every iteration and therefore all parallel instances are aware of the best solution and can share pheromones accordingly. High level pseudo code of such Parallel Ant (PA) implementation is shown in Figure 5. Main advantage of this architecture is that it allows efficient pheromone sharing, therefore converging faster. However, there is a high risk of getting stuck into local optima as all ants start iteration with the same pheromone matrix. Furthermore, synchronization of all parallel instances after every iteration is costly.

**Parallel Ants (PA)**

1. **for** all iterations *i* **do**
2.     **for** all parallel instances *m* **parallel do**
3.         **for** all local ants *a* **do**
4.            local pheromone = global pheromone
5.            construct solution
6.            local pheromone update
7.         **end for**
8.         find best solution across parallel instances
9.         update global pheromone update based on best solution
10.     **end for**
11. **end for**

*Figure 5. High level pseudo code Parallel Ants (PA) search algorithm*

Both IAC and PA implementations are exploiting task parallelism – each parallel instance (thread) gets set of tasks to complete. An alternative approach would be to look at data parallelism and vectorization – each thread processes a specific section of the data and cooperatively complete the given task. Due to the highly sequential parts of ACO, it would not be practical to only use vectorization alone. A more desirable path would be to implement vectorization in conjugate to the task parallelism. In case of CPU, task parallelism can be done by the threads, while vectorization done by Vector Processing Units (VPUs) based on Advanced Vector Extensions 2 (AVX2) or AVX512. Moreover, in case of GPU and CUDA – task parallelism would be done at thread-block level while data parallelism would exploit WARP structures. Parallel Ants with Vectorization (PAwV) expands on the Parallel Ants architecture by introducing data-parallelism of solution creation and an alternative roulette wheel implementation – SS-Wheel, first proposed in [24]. SS-Wheel mimics a sequential roulette wheel while allowing higher throughput due to parallelism. Local search in Figure 6 expands on the implementation in Figure 5 (lines 3-7). First the *choiceMatrix* is calculated by multiplying the probability of the route to be chosen with the *tabu list* – a list of still available routes (where 0 represents not available and 1 – route still can be selected). A random number between 0 and 1 is generated in order to determine if a given route will be chosen based on exploitation or exploration. In case of exploitation, the *choiceMatrix* is reduced to obtain the maximum and the corresponding route index. Furthermore, in case of exploration, the route is chosen based on the SS-Roulette wheel described by [24].

**Parallel Ants with Vectorization (PAwV)**

1. **for** all local ants *a* **do**
2.    local pheromone = global pheromone
3.    **for** all orders *o* **do**
4.       **for** all routes *r* for order **do SIMD**
5.          choiceMatrix[r] = probability[r] * tabuList[r]
6.       **end for**
7.       **if** rand() <= q0 **then**
8.          SIMD reduce max (choiceMatrix)
9.       **else**
10.         SS-Roulette wheel [24]
11.       **end if**
12.    **end for**
13.    local pheromone update
14. **end for**

*Figure 6. High level pseudo code for Parallel Ants with Vectorization (PAwV) search algorithm. Expanding on Figure 5 lines 3-7.*

## 6. Experiments

A sequential implementation of ACO described in [58] is adapted from [59] by altering the heuristic information calculation for a given route – defined as a proportion of order's weight and the maximum weight gap (see Equation (2)). Furthermore, the Ant Colony System set of parameters for all configurations and architectures are shown in Table 3. Moreover, we then implement three different Parallel ACO architectures – Independent Ant Colonies (IAC), Parallel Ants (PA) and Parallel Ants with Vectorization (PAwV) in C++ and CUDA C.

Experiments were conducted on three different hardware configurations – CPU, GPU and Xeon Phi. Where Hardware A is a host system for Hardware C.

*Table 3. Ant Colony System set of parameters for all configurations and architectures*

| Parameter | Value |
|---|---|
| Pheromone evaporation rate (rho) | 0.1 |
| Weight on pheromone information ($\alpha$) | 1 |
| Weight on heuristic information ($\beta$) | 8 |
| Exploitation to exploration ratio (q0) | 0.9 |

Hardware A - *CPU*

- CPU: AMD Ryzen™ Threadripper™ 1950X (16 cores, 32 threads), running at 3.85GHz.
- RAM: 64GB 2400MHz DDR4, 4 channels.
- OS: Windows 10 Pro, version 1703
- Toolchain: Intel C++ 18.0 toolset, Windows SDK version 8.1, x64

Hardware B - *Xeon Phi*

- CPU: Intel® Xeon Phi™ Processor 7250F (68 cores, 272 hyper-threads), running at 1.4GHz. Clustering mode set to *Quadrant* and memory mode set to *Cache mode*.
- RAM: 16GB on-chip MCDRAM and 96GB 2400MHz DDR4 ECC.
- OS: Windows Server 2016, version 1607
- Toolchain: Intel C++ 18.0 toolset, Windows SDK version 8.1, x64, KMP_AFFINITY=scatter

Hardware C - *GPU*

- CPU/RAM/OS – see host Hardware A.
- GPUs: 4x Nvidia GTX1070, 8GB GDDR5 per GPU, 1.9GHz core, 4.1GHz memory. PCIe with 16x/8x/16x/8x.
- Toolchain: Visual Studio v140 toolset, Windows SDK version 8.1, x64, CUDA 9.0, compute_35, sm_35

## 6.1. Benchmarks

It is important to take both elapsed time and solution quality into consideration when referring to speed optimization of optimization algorithms. One could get superior convergence within iteration but, take twice as long to compute. Similarly, one could claim that algorithm is much faster completing defined number of iterations, but sacrifice solution quality. Furthermore, there is little point comparing sequential execution of one hardware platform to a parallel implementation of another. Comparison should take into consideration all platform strengths and weaknesses and set up the most suitable configuration for given platform.

In order to obtain a baseline fitness convergence rate at various number of parallel instances, we create Iterations vs Parallel Instances matrix for all architectures. An example of such matrix for Parallel Ants is shown in Table 4. The matrix is derived by averaging the resulting fitness obtained from 10 independent simulations with a unique seed value for each given Parallel Instances configuration. All configurations are run for *x* number of iterations, where *x* is based on the total number of solutions explored and is a function of the number of Parallel Instances. The total number of solutions explored is set to 768k. The number of Parallel Instances is varied by $2^{n-1}$ with maximum n of 11, i.e. 1024 parallel instances. The best value after every 5 iterations is also recorded.

*Table 4. Parallel Ants fitness value baseline for different configurations of the number of parallel instances and the number of iterations. Each Parallel Instance datapoint is an average of 10 individual runs (table*

*derived from 11\*10 =110 runs). Expressed as a percentage of proximity of the best-known solution (2,701,367.58). Color coded from worse – in red, to the best – in green.*

**Baseline for Parallel Ants (PA)**

The number of Parallel Instances

| Iterations | 1 | 2 | 4 | 8 | 16 | 32 | 64 | 128 | 256 | 512 | 1024 |
|---|---|---|---|---|---|---|---|---|---|---|---|
| 5 | 98.646% | 98.701% | 98.740% | 98.713% | 98.813% | 98.825% | 98.857% | 98.859% | 98.881% | 98.931% | 98.923% |
| 20 | 98.921% | 98.935% | 98.973% | 98.987% | 98.980% | 99.063% | 99.053% | 99.082% | 99.102% | 99.133% | 99.150% |
| 40 | 99.165% | 99.265% | 99.315% | 99.300% | 99.343% | 99.355% | 99.366% | 99.413% | 99.410% | 99.427% | 99.443% |
| 60 | 99.354% | 99.413% | 99.466% | 99.503% | 99.530% | 99.536% | 99.541% | 99.562% | 99.573% | 99.592% | 99.598% |
| 80 | 99.438% | 99.459% | 99.547% | 99.547% | 99.585% | 99.585% | 99.582% | 99.630% | 99.638% | 99.660% | 99.667% |
| 100 | 99.444% | 99.459% | 99.548% | 99.559% | 99.589% | 99.592% | 99.584% | 99.646% | 99.641% | 99.672% | 99.674% |
| 200 | 99.452% | 99.461% | 99.551% | 99.569% | 99.601% | 99.605% | 99.599% | 99.724% | 99.717% | 99.846% | 99.844% |
| 300 | 99.452% | 99.461% | 99.558% | 99.574% | 99.615% | 99.615% | 99.606% | 99.734% | 99.743% | 99.869% | 99.878% |
| 400 | 99.456% | 99.464% | 99.559% | 99.577% | 99.615% | 99.628% | 99.631% | 99.739% | 99.763% | 99.877% | 99.885% |
| 500 | 99.456% | 99.465% | 99.560% | 99.584% | 99.624% | 99.637% | 99.641% | 99.739% | 99.772% | 99.884% | 99.891% |
| 600 | 99.456% | 99.471% | 99.560% | 99.584% | 99.624% | 99.641% | 99.643% | 99.740% | 99.772% | 99.891% | 99.898% |
| 750 | 99.458% | 99.474% | 99.560% | 99.588% | 99.634% | 99.647% | 99.645% | 99.753% | 99.778% | 99.896% | 99.901% |
| 1500 | 99.462% | 99.494% | 99.572% | 99.590% | 99.638% | 99.662% | 99.656% | 99.764% | 99.792% | 99.917% | |
| 3000 | 99.471% | 99.504% | 99.582% | 99.601% | 99.651% | 99.672% | 99.666% | 99.779% | 99.812% | | |
| 6000 | 99.486% | 99.506% | 99.596% | 99.616% | 99.659% | 99.675% | 99.675% | 99.787% | | | |
| 12000 | 99.494% | 99.517% | 99.604% | 99.626% | 99.666% | 99.681% | 99.692% | | | | |
| 24000 | 99.498% | 99.540% | 99.611% | 99.629% | 99.681% | 99.693% | | | | | |
| 48000 | 99.508% | 99.546% | 99.622% | 99.638% | 99.685% | | | | | | |
| 96000 | 99.514% | 99.555% | 99.622% | 99.643% | | | | | | | |
| 192000 | 99.527% | 99.563% | 99.622% | | | | | | | | |
| 384000 | 99.538% | 99.569% | | | | | | | | | |
| 768000 | 99.551% | | | | | | | | | | |

We then compute the number of iterations required to reach a specific solution quality for different ACO architectures in Table 5, expressed as proximity to the best-known optimal solution. For the specific problem and dataset, the best solution is a total cost of 2,701,367.58. There are 6 checkpoints of solution quality ranging from 99% to 99.9%. Although at first 1% gain might not seem significant, one has to remember that global supply chain costs are measured in hundreds of millions, and even 1% savings do affect bottom line. Empty fields (-) represent instances where the ACO was not able to converge to given solution quality.

On all experiments, IAC was able to obtain solution quality only below 99.6%, whereas PA and PA with 5 ant local search was able to obtain optimal solution with 512 and 1024 parallel instances. Furthermore, IAC did not see any significant benefit of adding more parallel instances for 99% and 99.25% checkpoints.

In contrast, PA does benefit from the increase in number of parallel instances. For instance, PA is able to obtain the same solution quality in half the number of iterations at 99% checkpoint (scaling of 2x for sequential vs 1024 parallel instances). Scaling of 633.7x in case of 99.5% checkpoint for sequential counterpart. Similarly, PA with 5 ant sequential local search has the same dynamics, with scaling of 4x at 99% checkpoint compared to sequential and 140x at 99.6% checkpoint compared to 2 and 1024 parallel instances. One can also note that at increased solution quality and little number of parallel instances, PA with 5 ant local search also offers increased efficiency in terms of total solutions explored. For example, at the 99.5% checkpoint with 2 parallel instances, PA takes 2590 iterations, while PA with 5 ant local search only requires 65 (decrease of 40x iterations, or 8x total solutions explored). However, in most instances, PA without any local search is more efficient.

*Table 5. The number of iterations required to reach a specific solution quality. Each datapoint in the table is an average of 10 individual runs. Empty fields (-) represent instances where ACO did not obtain specified solution quality in 768k solutions explored. The solution quality is expressed as a percentage of proximity of the best-know solution (2,701,367.58).*

| Architecture | Checkpoint of optimal solution | The number of iterations required to reach specific solution quality |||||||||||
| --- | --- | --- | --- | --- | --- | --- | --- | --- | --- | --- | --- |
| | | The number of parallel instances |||||||||||
| | | 1 | 2 | 4 | 8 | 16 | 32 | 64 | 128 | 256 | 512 | 1024 |
| Independent Ant Colonies | 99.00% | 30 | 30 | 35 | 30 | 30 | 35 | 30 | 30 | 25 | 25 | 25 |
| | 99.25% | 45 | 45 | 40 | 40 | 45 | 40 | 40 | 35 | 35 | 35 | 35 |
| | 99.50% | 31685 | 31055 | 29550 | 28895 | 29075 | 15910 | 10950 | - | - | - | - |
| | 99.60% | - | - | - | - | - | - | - | - | - | - | - |
| | 99.75% | - | - | - | - | - | - | - | - | - | - | - |
| | 99.90% | - | - | - | - | - | - | - | - | - | - | - |
| Parallel Ants | 99.00% | 30 | 25 | 25 | 25 | 25 | 25 | 20 | 15 | 15 | 15 | 15 |
| | 99.25% | 45 | 40 | 40 | 35 | 35 | 35 | 35 | 35 | 30 | 30 | 30 |
| | 99.50% | 31685 | 2590 | 65 | 60 | 60 | 55 | 55 | 55 | 55 | 50 | 50 |
| | 99.60% | - | - | 9190 | 2640 | 195 | 170 | 230 | 70 | 70 | 65 | 65 |
| | 99.75% | - | - | - | - | - | - | - | 685 | 310 | 140 | 135 |
| | 99.90% | - | - | - | - | - | - | - | - | - | 800 | 675 |
| Parallel Ants with 5 sequential ant local search | 99.00% | 20 | 15 | 15 | 15 | 15 | 10 | 10 | 10 | 10 | 10 | 5 |
| | 99.25% | 30 | 30 | 30 | 30 | 30 | 25 | 30 | 25 | 20 | 25 | 20 |
| | 99.50% | 400 | 65 | 55 | 55 | 50 | 50 | 50 | 50 | 45 | 45 | 45 |
| | 99.60% | - | 7715 | 160 | 135 | 90 | 65 | 60 | 65 | 60 | 55 | 55 |
| | 99.75% | - | - | - | - | 6630 | 205 | 150 | 155 | 130 | 125 | 125 |
| | 99.90% | - | - | - | - | - | - | - | - | 460 | 255 | 160 |

## 6.2. Speed performance

To evaluate speed performance, we ran each given configuration and parallel architecture for 500 iterations or 10 minutes wall-clock time (whichever happens first) and recorded total number of iterations and wall-clock time for 3 independent runs. Then, average wall-clock time per iteration was calculated. It is important to measure the execution time correctly, just purely comparing computation per kernel/method may not show the real-life impact. For that reason, total time is measured from the start of the memory allocation to the freeing of the allocated memory, however it does not include time required to load the dataset into memory. This allows us to estimate, with reasonable accuracy, what is the wall-clock time required to run a specific architecture and configuration in order to converge to a given fitness quality. Although, running each given architecture and configuration 10 times would produce more accurate convergence rate estimates, it would also require significantly more computation time. Furthermore, all vectorized implementations went through iterative profiling and optimization process to obtain the fastest execution time. To the best of the authors' knowledge, all vectorized implementations have been fully optimized for the given hardware.

### 6.2.1. CPU

ACO implementation of IAC, PA and PAwV was implemented in C++ and multiple experiments of the configurations are shown in Table 6. Intel C++ 18.0 with OpenMP 4.0 was used to compile the implementation. KMP[1] (an extension of OpenMP) config was varied based on total hardware core and logical core count (16c,2t = 32 OpenMP threads).

Very similar results were obtained for both IAC double precision and PA double precision, with PA having around 5% overhead compared to IAC. In both instances, running 32 OpenMP threads offered around 24% speed reduction compared to 16 threads. Furthermore, PAwV with double precision vectorization using AVX2 offered speed reduction of 26%, while scaling from 16 OpenMP threads to 32 offered almost no scaling at 256 parallel instances upwards.

The nature of ACO pheromone sharing and probability calculations does not require double precision and therefore can be substituted with single precision calculations. AVX2 offers 256-bit manipulations, therefore increasing theoretical throughput by factor of 2, compared to double precision. 36% decrease in execution time was obtained, as not all parts of the code are able to take advantage of SIMD.

Furthermore, doing 5 ant sequential local search within each parallel instance increases time linearly and produces little time savings in terms of solutions explored. The overall scaling factor at 1024 parallel instances compared to sequential execution at PAwV (single precision with AVX2 and 16c2t) is therefore 25.4x.

*Table 6. Hardware A wall-clock time per iteration, in seconds. KMP config is environment variable set as part of KMP_PLACE_THREADS, for all instances KMP_AFFINITY=scatter, optimization level /O3, favor speed /Ot.*

| Hardware A - CPU computation time per iteration (in seconds) | | | | | | | | | | | | |
|---|---|---|---|---|---|---|---|---|---|---|---|---|
| Configuration | | The number of Parallel Instances | | | | | | | | | | |
| | KMP config | 1 | 2 | 4 | 8 | 16 | 32 | 64 | 128 | 256 | 512 | 1024 |
| IAC, double precision | 16c,1t | 0.078 | 0.081 | 0.083 | 0.085 | 0.112 | 0.196 | 0.372 | 0.691 | 1.368 | 2.661 | 5.263 |
| | 16c,2t | | | | | | 0.148 | 0.277 | 0.517 | 1.002 | 2.014 | 4.093 |
| PA, double precision | 16c,1t | 0.082 | 0.084 | 0.085 | 0.090 | 0.115 | 0.205 | 0.383 | 0.705 | 1.411 | 2.743 | 5.483 |
| | 16c,2t | | | | | | 0.153 | 0.288 | 0.539 | 1.044 | 2.088 | 4.220 |
| PAwV, double precision, AVX2 | 16c,1t | 0.050 | 0.053 | 0.057 | 0.058 | 0.075 | 0.131 | 0.233 | 0.426 | 0.805 | 1.547 | 3.101 |
| | 16c,2t | | | | | | 0.107 | 0.189 | 0.351 | 0.749 | 1.536 | 3.095 |
| PAwV, single precision, AVX2 | 16c,1t | 0.049 | 0.050 | 0.052 | 0.055 | 0.066 | 0.111 | 0.206 | 0.367 | 0.699 | 1.355 | 2.664 |
| | 16c,2t | | | | | | 0.088 | 0.152 | 0.275 | 0.501 | 1.006 | 1.975 |
| PAwV, single precision, AVX2, with 5 sequential ant local search | 16c,1t | 0.212 | 0.218 | 0.227 | 0.241 | 0.264 | 0.484 | 0.918 | 1.722 | 3.380 | 6.759 | 13.461 |
| | 16c,2t | | | | | | 0.347 | 0.645 | 1.222 | 2.369 | 4.659 | 9.704 |

### 6.2.2. Xeon Phi

Similar experiments were conducted also on the Xeon Phi hardware, Table 7. Due to the poor convergence rate and search capability, execution time for IAC was not measured. Xeon Phi differs from Hardware A with the ability to utilize up to 4 hyper-threads per core and AVX512 instruction set. Although Hardware B has 68 physical cores, for simpler comparison on base 2, only 64 were used in experiments. At 1024 parallel instances on double precision PA, having 2 threads and 4 threads per core does offer speedup of 30% and 42% respectively, compared to 1 thread per core. Moving to the vectorized implementation of 256-bit AVX2, gains additional speedup of around 37% across all parallel instances, however, did not benefit from 4 hyper-threads. Furthermore, exploiting the AVX512 instruction set offers further 24% speedup compared to AVX2. In this configuration having 4 hyper threads per core actually worsens the speed performance (3.644 seconds vs 3 seconds). Similar to Hardware A, PAwV was explored with single precision and offered near perfect scaling on 1024 parallel instances with 4 hyper-threads per core, or 40% overall speed improvement compared to PAwV with double precision (3 seconds vs 1.804 seconds). Alike Hardware A, having 5 sequential local ants does not provide any time savings and time increases linearly. The overall scaling factor at 1024 parallel instances compared to sequential execution at PAwV (single precision with AVX512 and 64c4t) is therefore 148x.

---
[1] OpenMP Thread Affinity Control https://software.intel.com/en-us/articles/openmp-thread-affinity-control

*Table 7. Hardware B wall-clock time per iteration, in seconds. KMP config is environment variable set as part of KM_PLACE_THREADS, for all instances KMP_AFFINITY=scatter, optimization level /O3, favor speed /Ot.*

| Hardware B - Xeon Phi computation time per iteration (in seconds) | | | | | | | | | | | | |
|---|---|---|---|---|---|---|---|---|---|---|---|---|
| Configuration | | The number of Parallel Instances | | | | | | | | | | |
| | KMP config | 1 | 2 | 4 | 8 | 16 | 32 | 64 | 128 | 256 | 512 | 1024 |
| PA, double precision | 64c,1t | | | | | | | | 1.417 | 2.787 | 5.941 | 11.089 |
| | 64c,2t | 0.687 | 0.687 | 0.725 | 0.726 | 0.726 | 0.729 | 0.734 | 1.014 | 1.974 | 3.845 | 7.669 |
| | 64c,4t | | | | | | | | 1.087 | 1.606 | 3.226 | 6.438 |
| PAwV, double precision, AVX2 | 64c,1t | | | | | | | | 0.818 | 1.578 | 3.094 | 6.114 |
| | 64c,2t | 0.408 | 0.411 | 0.430 | 0.431 | 0.433 | 0.434 | 0.438 | 0.563 | 1.047 | 2.022 | 3.964 |
| | 64c,4t | | | | | | | | 0.625 | 1.101 | 2.072 | 4.082 |
| PAwV, double precision, AVX512 | 64c,1t | | | | | | | | 0.608 | 1.152 | 2.242 | 4.404 |
| | 64c,2t | 0.304 | 0.309 | 0.326 | 0.326 | 0.327 | 0.332 | 0.335 | 0.446 | 0.809 | 1.535 | 3.000 |
| | 64c,4t | | | | | | | | 0.494 | 0.982 | 1.913 | 3.644 |
| PAwV, single precision, AVX512 | 64c,1t | | | | | | | | 0.521 | 0.970 | 1.900 | 3.806 |
| | 64c,2t | 0.261 | 0.266 | 0.282 | 0.284 | 0.284 | 0.287 | 0.288 | 0.359 | 0.646 | 1.210 | 2.361 |
| | 64c,4t | | | | | | | | 0.412 | 0.542 | 0.957 | 1.804 |
| PAwV, single precision, AVX512, with 5 sequential ant local search | 64c,1t | | | | | | | | 2.342 | 4.601 | 9.136 | 18.844 |
| | 64c,2t | 1.105 | 1.123 | 1.195 | 1.200 | 1.205 | 1.205 | 1.215 | 1.489 | 2.915 | 5.743 | 11.815 |
| | 64c,4t | | | | | | | | 1.553 | 2.225 | 4.428 | 9.054 |

### 6.2.3. GPUs

A further set of experiments were also conducted for GPU, Table 8. The implementation with no vectorization (Blocks x1), uses 1 thread per CUDA block to compute one solution, therefore 1024 parallel instances require 1024 blocks. Similarly, for (Blocks x32), 32 threads are used per block, each thread computing its own solution independently. For parallel instances of 32, only 1 block would be used with 32 threads. The implementation of no vectorization utilizes no shared memory, however, all static problem meta data is stored as textures. A single kernel is launched and best solution across all parallel instances is returned.

Vectorized version implements architecture described in [24], storing the route choice matrix in shared memory and utilizing local warp reduction for sum and max operations. Each thread-block builds its own solution, while the extra 32 threads assist with the reduction operations, memory copies and fitness evaluation. Table 8 shows the comparison between the two implementations. Implementation without vectorization performs on average 2 times slower compared to the vectorized version. Furthermore, 64 threads per block (Blocks x64) performs slower than 32 threads per block (Block x32).

Next, scaling across multiple GPUs were explored. Each device takes a proportion of 1024 instances with unique seed values and after each iteration, best overall solution is reduced. In case of 2 GPUs and 1024 parallel instances, each device will compute 512 parallel instances concurrently. Scaling across 2 (2x) and 4 GPUs (4x) did not provide any significant speedup (only 10%). This is due to the fact that each iteration consumes at least 50 seconds and scaling across multiple GPUs adds almost no overhead. The maximum number of parallel instances might need to be increased to fully utilize all 4 GPUs to the point where all Streaming Multiprocessors (SMs) are saturated and increasing block count increases the computation time linearly.

GPU implementation is therefore one magnitude of order slower than that of CPU, though this could be explained by the nature of the problem and not be specific to ACO architecture, as there have been a lot of success on GPUs solving simple, low memory footprint TSP instances [24] [46] [47]. However, the problem being solved in this paper requires a lot of random global memory access to check for all restrictions such as order limits, capacity constraints and weight limits, which are too big to be stored on the shared memory.

*Table 8. Hardware C wall-clock time per iteration, in seconds. Total number of parallel instances are adjusted for the thread-block dimensions. Compiled with CUDA 9.0. 1x, 2x and 4x correspond of number of devices used to compute.*

| Configuration | Hardware C - GPU computation time per iteration (in seconds) | | | | | | | | | | |
|---|---|---|---|---|---|---|---|---|---|---|---|
| | The number of Parallel Instances | | | | | | | | | | |
| | 1 | 2 | 4 | 8 | 16 | 32 | 64 | 128 | 256 | 512 | 1024 |
| 1x GPU no vectorisation (Blocks x 1) | 46.792 | 47.634 | 47.610 | 47.499 | 47.458 | 48.914 | 50.811 | 53.474 | 60.845 | 126.897 | 229.080 |
| 1x GPU no vectorisation (Blocks x 32) | - | - | - | - | - | 108.316 | 110.571 | 112.512 | 113.214 | 114.512 | 115.219 |
| 1x GPU with vectorisation (Blocks x32) | - | - | - | - | - | 49.890 | 52.457 | 54.180 | 55.409 | 58.802 | 64.569 |
| 1x GPU with vectorisation (Blocks x64) | - | - | - | - | - | - | 57.139 | 58.586 | 59.676 | 61.031 | 65.840 |
| 2x GPU with vectorisation (Blocks x32) | - | - | - | - | - | - | 50.048 | 52.634 | 55.471 | 55.509 | 60.856 |
| 4x GPU with vectorisation (Blocks x32) | - | - | - | - | - | - | - | 50.062 | 52.702 | 54.406 | 55.879 |

## 6.3. Hardware Comparison and speed of convergence

If both convergence rate of the architecture and the speed of the hardware is taken into account, an estimate can be made on what would be the average wall-clock time to converge to a specific solution quality. The fastest configuration for both Hardware A (Table 6) and Hardware B (Table 7) was chosen and then multiplied by the number of iterations required to reach a specific solution quality (Table 5) to obtain an estimate of the compute time required (Table 9). Therefore, a fairer real-life impact can be derived. GPU results (Hardware C) were not included as they are significantly slower.

If one only considers the best time to converge to 99% solution quality, Hardware A can do that in 1.24 seconds on average while, Hardware B would take 6.66 seconds. Furthermore, if we look at 99.5% solution quality, Hardware A would take 3.33 seconds while Hardware B - 17.01 seconds. Faster clock speed for Hardware A gives advantage over Hardware B at lower solution quality checkpoints. In contrast, at 99.75% and 99.9% solution quality, Hardware B outperforms. More experimentation is required to determine if exploring more than 768k solutions at lower Parallel Instance count affects the dynamics at the 99.75-99.9% range.

Table 9. *Estimated time (in seconds) required to converge to a specific solution quality. Calculated by multiplying the number of iterations by the time taken for iteration for individual best performing hardware configuration. Solution quality is expressed as a percentage of proximity of the best-know solution (2,701,367.58).*

| Architecture | Checkpoint of optimal | Estimated time required (in seconds) to reach specific solution quality | | | | | | | | | | |
|---|---|---|---|---|---|---|---|---|---|---|---|---|
| | | The number of parallel instances | | | | | | | | | | |
| | | 1 | 2 | 4 | 8 | 16 | 32 | 64 | 128 | 256 | 512 | 1024 |
| Hardware A - TR1950x | 99.00% | 1.46 | 1.24 | 1.30 | 1.39 | 1.64 | 2.19 | 3.04 | 4.13 | 7.52 | 15.10 | 29.63 |
| | 99.25% | 2.19 | 1.99 | 2.07 | 1.94 | 2.29 | 3.06 | 5.31 | 9.64 | 15.03 | 30.19 | 59.25 |
| | 99.50% | 1539.02 | 128.82 | 3.37 | 3.33 | 3.93 | 4.81 | 8.35 | 15.14 | 27.56 | 50.32 | 98.75 |
| | 99.60% | | | 476.40 | 146.33 | 12.78 | 14.88 | 34.92 | 19.27 | 35.07 | 65.42 | 128.38 |
| | 99.75% | | | | | | | | 188.60 | 155.33 | 140.91 | 266.63 |
| | 99.90% | | | | | | | | | | 805.20 | 1333.13 |
| Hardware B - Xeon Phi 7250F | 99.00% | 7.84 | 6.66 | 7.04 | 7.09 | 7.10 | 7.18 | 5.76 | 6.18 | 8.13 | 14.36 | 27.06 |
| | 99.25% | 11.76 | 10.65 | 11.27 | 9.92 | 9.94 | 10.05 | 10.08 | 14.42 | 16.26 | 28.71 | 54.12 |
| | 99.50% | 8282.30 | 689.67 | 18.31 | 17.01 | 17.04 | 15.79 | 15.84 | 22.66 | 29.81 | 47.85 | 90.20 |
| | 99.60% | | | 2588.73 | 748.49 | 55.39 | 48.80 | 66.26 | 28.84 | 37.94 | 62.21 | 117.26 |
| | 99.75% | | | | | | | | 282.22 | 168.02 | 133.98 | 243.54 |
| | 99.90% | | | | | | | | | | 765.60 | 1217.70 |

## 7. Conclusions & Further work

Nature-inspired meta-heuristic algorithms such as Ant Colony Optimization (ACO) have been successfully applied to multiple different optimization problems. Most work focuses on the Travelling Salesman Problem (TSP). While TSPs are a good benchmark for new idea comparison, the dynamics of the proposed algorithms for benchmarks do not always match to a real-world performance where problem has more constraints (more meta-data during solution creation). Furthermore, speed and fitness performance comparisons are not always completely fair when compared to a sequential implementation.

This work solves a real-world outbound supply chain network optimization problem and compares two different ACO architectures – Independent Ant Colonies (IAC) and Parallel Ants (PA). It was concluded that PA outperformed IAC in all instances, as IAC failed to find any better solution than 99.5% of optimal. In comparison, PA was able to find near optimal solution (99.9%) in less iterations due to efficient pheromone sharing across ants after each iteration. Furthermore, PA shows that it consistently finds a better solution with the same number of iterations as the number of parallel instances increase.

Moreover, a detailed speed performance was measured for 3 different hardware architectures – 16 core 32 thread workstation CPU, 68 core server grade Xeon Phi and general purpose Nvidia GPUs. Results showed that due to the nature of the real-world problem, memory access footprint required to check capacity limits and weight constraints did not fit on small shared memory on GPU and therefore it performed 29 times slower than the other two hardware solutions even when running 4 GPUs in parallel.

When compared to a real-life impact on time required to reach a specific solution quality, both CPU and Xeon Phi optimized-vectorized implementations showed comparable speed performance; with CPU taking the lead with lower Parallel Instances count due to the much higher clock frequency. At near optimal solution (99.75%+) and 1024 parallel instances, Xeon Phi was able to take full advantage of AVX512 instruction set and outperformed CPU in terms of speed. Therefore, compared to an equivalent sequential implementation at 1024 parallel instances, CPU was able to scale 25.4x while Xeon Phi achieved a speedup of 148x.

Due to the fact that PA fitness performance increases as the number of parallel instances increase, it would be worth looking into scaling above 1024 instances using either clusters of CPUs or clusters of Xeon Phi's, which will be part of the future work. Furthermore, Field Programmable Gate Arrays (FPGAs) might have potential to take advantage of highly vectorized ACO, which is another area of possible future research.

**Acknowledgment**

Authors would like to thank Intel Corporation for donating the Xeon Phi hardware.